\documentclass[letterpaper, 10 pt, conference]{ieeeconf}  % Comment this line out if you need a4paper

\IEEEoverridecommandlockouts                              % This command is only needed if 
\usepackage{graphics} % for pdf, bitmapped graphics files
\usepackage{epsfig} % for postscript graphics files
\usepackage{times} % assumes new font selection scheme installed
\usepackage{amsmath} % assumes amsmath package installed
\usepackage{amssymb}  % assumes amsmath package installed

\usepackage[T1]{fontenc}
\usepackage{subfigure}
\usepackage{bm}
\usepackage{diagbox}

\usepackage{multirow} %合并多行单元格的宏包
\usepackage{longtable} %不宽但很长的表格可以用longtable宏包来进行分页显示
\usepackage{array} %一般用于数学公式中对数组或矩阵的排版
\usepackage{makecell}
\usepackage{bbding}% makecell命令对表格单元格中的数据进行一些变换的控制。我们可以使用 \ 命令进行换行，也可以使用p{(宽度)}选项控制列表的宽度
\usepackage{threeparttable} %制作三线表格
\usepackage{booktabs}%s三线表格中的上中下直线线型设置宏包，在这个包中水平线被教程\toprule、midrule、buttomrule。
\usepackage[colorlinks,linkcolor=blue]{hyperref}

\title{\LARGE \bf
DRKF: Distilled Rotated Kernel Fusion for Efficient Rotation Invariant Descriptors in Local Feature Matching
}

\author{Ranran Huang$^{1}$, Jiancheng Cai$^{1}$, Chao Li$^{2}$, Zhuoyuan Wu$^{1}$, Xinmin Liu$^{1}$, Zhenhua Chai$^{1}$% <-this % stops a space
\thanks{$^{1}$Ranran Huang, Jiancheng Cai, Zhuoyuan Wu, Xinmin Liu and Zhenhua Chai are with Meituan, No.7 Rongda Road, Chaoyang District, Beijing, 100012, China
        {\tt\small \{huangranran, caijiancheng, wuzhuoyuan02, liuxinmin, chaizhenhua\}@meitua n.com}}
\thanks{$^{2}$Chao Li is with School of Artificial Intelligence, Beijing University of Posts and Telecommunications, No.10 Xitucheng Road, Haidian District, Beijing, 100876, China
        {\tt\small chaoli@bupt.edu.cn}}
}

\begin{document}

\maketitle
\thispagestyle{empty}
\pagestyle{empty}

%%%%%%%%%%%%%%%%%%%%%%%%%%%%%%%%%%%%%%%%%%%%%%%%%%%%%%%%%%%%%%%%%%%%%%%%%%%%%%%%
\begin{abstract}

% Most existing learning-based image matching pipelines are designed for better feature detectors and descriptors which are robust to repeated textures, viewpoint changes, etc., while little attention has been paid to rotation invariance. As a consequence, these approaches usually demonstrate inferior performance compared to the handcrafted algorithms in circumstances where a significant level of rotation exists in data, due to the lack of keypoint orientation prediction. 

The performance of local feature descriptors degrades in the presence of large rotation variations.
To address this issue, we present an efficient approach to learning rotation invariant descriptors.
Specifically, we propose Rotated Kernel Fusion (RKF) which imposes rotations on the convolution kernel to improve the inherent nature of CNN.
Since RKF can be processed by the subsequent re-parameterization, no extra computational costs will be introduced in the inference stage.
Moreover, we present Multi-oriented Feature Aggregation (MOFA) which aggregates features extracted from multiple rotated versions of the input image and can provide auxiliary knowledge for the training of RKF by leveraging the distillation strategy. We refer to the distilled RKF model as DRKF. 
Besides the evaluation on a rotation-augmented version of the public dataset HPatches, we also contribute a new dataset named DiverseBEV which is collected during the drone's flight and consists of bird's eye view images with large viewpoint changes and camera rotations.
Extensive experiments show that our method can outperform other state-of-the-art techniques when exposed to large rotation variations.
\end{abstract}

%%%%%%%%%%%%%%%%%%%%%%%%%%%%%%%%%%%%%%%%%%%%%%%%%%%%%%%%%%%%%%%%%%%%%%%%%%%%%%%%
\section{INTRODUCTION}

Local feature matching between images is essential for multiple computer vision tasks, including Structure from Motion (SfM) \cite{schonberger2016structure} and Simultaneous Localization and Mapping (SLAM) \cite{mur2015orb}.
In recent years, this field has obtained significant development due to the advance in deep learning,
however, 
it is still challenging to 
find reliable corresponding points across images with large geometric transformations and appearance differences caused by camera rotations.
% For instance, drone-mounted cameras captures the scene image with a wide angle.
% therefore, the local image descriptors are required to be invariant to geometric transformations caused by camera rotations, 

% especially under circumstances where cameras move in 3D space and severe camera rotations and viewpoint changes are commonplace, such as in matching task for bird's eye view images. 

Traditional local descriptors including SIFT \cite{lowe2004distinctive}, SURF \cite{bay2008speeded} and ORB \cite{rublee2011orb} employ a detector-descriptor pipeline and 
explicitly assign orientations to each keypoint, thereby achieving rotation invariance by design. 
% However, they lack the ability to cope with the complex cases caused by repeated textures, viewpoint changes, etc. 
Recently, numerous learning-based methods have proved their potential in dealing with complicated cases. 
Methods like LIFT \cite{yi2016lift}, LF-Net \cite{ono2018lf}, and RF-Net \cite{shen2019rf} follow the idea of SIFT \cite{lowe2004distinctive} and predict the shape parameters 
% (including orientations and scales) 
based on which image patches around each keypoint are cropped and warped for further description. However, the process brings huge computational costs and fails to handle very large rotation changes due to the lack of explicit labels.
On the contrary, most popular learning-based implementations like SuperPoint \cite{detone2018superpoint}, D2-Net \cite{dusmanu2019d2}, R2D2 \cite{revaud2019r2d2} and ASLFeat \cite{luo2020aslfeat} directly locate the keypoints and compute their descriptors using CNN without considering orientations, therefore there exists a more significant performance gap in the presence of large rotation variations.
Although there are some methods aiming to obtain rotation invariant features \cite{cohen2016group,cohen2016steerable,bokman2022case,parihar2021rord,liu2019gift}, they increase the computational overhead and the difficulty with deployment in embedded devices.

% \begin{figure} 
% \centering    
% \subfigure[SIFT correspondences under rotation of $\pi$.] {
%  \label{rotinv-sift}     
% \includegraphics[width=0.33\textwidth]{sift-vis.jpg}  
% }     
% \subfigure[LF-Net correspondences under rotation of $\pi$.] { 
% \label{rotinv-lf}     
% \includegraphics[width=0.33\textwidth]{lfnet-vis.jpg}     
% }    
% \subfigure[ASLFeat correspondences under rotation of $\pi$.] { 
% \label{rotinv-asl}     
% \includegraphics[width=0.33\textwidth]{aslfeat-vis.jpg}     
% }
% \vspace{-5pt}
% \caption{Visualized matching results of SIFT, LF-Net and ASLFeat under rotation of $\pi$. The correspondences colored green are inliers while those colored red are outliers.}   

% \label{rotinv}     
% \end{figure}

\begin{figure*}[htbp]
    \centering
    \includegraphics[width=0.76\textwidth]{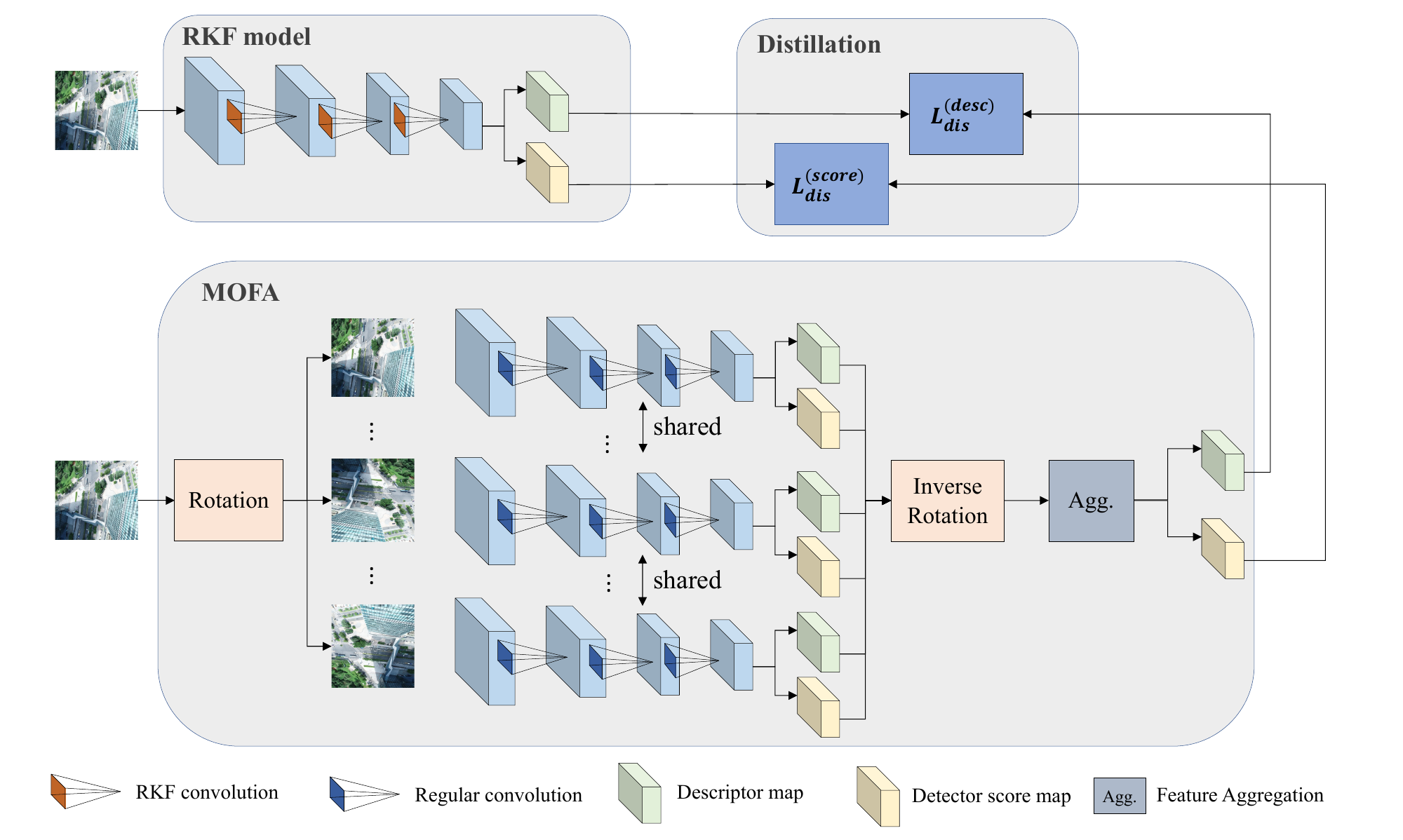}
    \vspace{-10pt}
    \caption{The figure shows the framework of our method. We follow the detection-and-description framework to jointly optimize the detection and description objectives. The RKF model replaces the regular convolution with RKF convolution and improves the inherent nature of CNN (see details in Fig. \ref{RKF}). MOFA integrates the features extracted from multiple rotated versions of images and serves as the teacher model to provide auxiliary supervision to the RKF model. The distillation stage generates the distilled RKF model (DRKF).}
    \vspace{-15pt}
    \label{pipeline}
\end{figure*}

In this paper, we address the challenge of large or even extreme rotation changes by making the following contributions to improve rotation invariance efficiently:
% \begin{itemize}
%     \item Instead of predicting orientations and scales as classical algorithms do, a model that aggregates features of multiple orientations is proposed. By doing this, learning rotation invariance for large orientation differences can be converted to learning that for multiple small orientation differences, which is beneficial for easing the representation burden of CNNs.
%     \item Knowledge distillation is the utilized, where the multi-oriented feature aggregation model (MOFA) aforementioned is regarded as teacher for rotation invariance transfer to the student model. Compared to the general case when the model is trained with data augmentation, distillation with MOFA can further improve rotation invariance.
%     \item To enhance learning capacity of the student model and improve knowledge transfer efficiency, the rotated kernel fusion (RKF) is applied in each convolution layer, where features of different orientations are obtained via rotating kernels and then merged using a learnable manner. Eventually, reparameterization is adopted, by which computational costs are not increased in inference stage.
% \end{itemize}
\begin{itemize}
    \item Shown in Fig. \ref{pipeline} and Fig. \ref{RKF}, we propose Rotated Kernel Fusion \textbf{(RKF)} to impose rotations on each convolution kernel followed by fusion on features extracted by multiple rotated filters. Employing RKF, we enable CNN to be rotation invariant inherently. Eventually, we leverage re-parameterization to make the computational cost remain the same in the inference stage without affecting the performance.
    \item By introducing multiple rotation transformations to the input images and conducting feature ensemble, we introduce Multi-oriented Feature Aggregation \textbf{(MOFA)} shown in Fig. \ref{pipeline}, which is further used as a teacher in the knowledge distillation pipeline to supervise the model equipped with RKF operation. 
    % To be specific, MOFA introduces multiple rotation transformations to the input images and ensembles the corresponding extracted features.  
    \item Due to the lack of publicly available aerial-view datasets with large viewpoint changes and camera rotations, we construct a new dataset named DiverseBEV which is collected during the drone's flight.
    % \item  Instead of directly transferring knowledge from MOFA to the base model, we propose to 
    % adopt the RKF model as the student to learn from MOFA, and the distilled model is referred to as Distilled Rotated Kernel Fusion (\textbf{DRKF}) which is shown in Fig. 2.
    \item Our experimental results show that DRKF improves rotation invariance considerably and outperforms other state-of-the-art approaches remarkably on rotation-augmented HPatches \cite{balntas2017hpatches} and the new DiverseBEV dataset without introducing extra computational cost in the inference phase. 
\end{itemize}

\section{Related Works}

\subsection{Local Feature Matching}
Hand-crafted methods including SIFT \cite{lowe2004distinctive}, SURF \cite{bay2008speeded} and ORB \cite{rublee2011orb} have dominated the field for a long time.
Many classical descriptors compute the orientations of detected keypoints, so that they can be rotation invariant by design. For example, SIFT \cite{lowe2004distinctive} first selects the keypoints using local extrema detection in the pyramid of Difference of Gaussian (DoG) and obtains orientations with the Histogram of Oriented Gradients (HOG). 
% When constructing descriptors, the coordinate axis for each keypoint is rotated according to its orientation. 
Some other methods obtain the orientation based on intensity. For instance, ORB \cite{rublee2011orb} first detects keypoints with FAST \cite{rosten2006machine} in different scales and then takes the direction from the corner's center to the intensity centroid in a local window as the orientation.
Despite the robustness to rotation, they lack the ability to cope with complex cases caused by weak or repeated textures and illumination changes.

Recently, we see major improvements in learning-based methods which offer promising advances in the capability of dealing with complicated cases.
In terms of rotation invariance, methods including LIFT \cite{yi2016lift}, LF-Net \cite{ono2018lf}, RF-Net \cite{shen2019rf} imitate the traditional methods and predict the orientation and scale parameters. The pipeline usually consists of feature detection, patch sampling with spatial transform network (STN) \cite{jaderberg2015spatial} and feature description. However, it is challenging to regress feature scales and orientations accurately without explicit labels. In addition, patch sampling and warping for numerous detected keypoints are computationally expensive.
% and overlapping areas among the patches also cause redundant computation.
On the contrary, recent techniques like SuperPoint \cite{detone2018superpoint}, D2-Net \cite{dusmanu2019d2} and ASLFeat \cite{luo2020aslfeat} achieve state-of-the-art results. For example, D2-Net and ASLFeat 
use a detect-and-describe approach, and couple the capability of the feature detector and descriptor by deriving keypoints from the same feature maps that are used for extracting feature descriptors.
However, they neglect the orientations of keypoints, therefore fail to demonstrate steady performance when exposed to extreme viewpoint variations.
The same problem still exists for recent local feature matching techniques like SuperGlue \cite{sarlin2020superglue} and LoFTR \cite{sun2021loftr}. 

To obtain transformation invariant descriptors, GIFT \cite{liu2019gift} uses group convolutions \cite{cohen2016group} to exploit underlying structures of the extracted features from the transformed versions of an image. RoRD \cite{parihar2021rord} is based on D2-Net \cite{dusmanu2019d2} and achieves rotation-robustness by training with in-plane rotated images and performing correspondence ensemble, however, its expressive capacity is still limited by the inherent nature of CNN. 
There are some other CNN-based alternatives developed for rotation-equivariance \cite{cohen2016group,cohen2016steerable,weiler2019general, bokman2022case,peri2022ref}, however, they require to re-design the CNN architectures, leading to increasing computational burden in the inference stage or difficulty with deployment in embedded devices.
Compared to these methods, our method 1) improves the inherent property of convolution layers without altering the architecture for inference; 2) aggregates the features extracted from rotated versions of input images just for auxiliary supervision in the training phase, which introduces no extra computational overhead for inference.

\subsection{Re-parameterization}

Generally, models with multi-branch aggregation usually demonstrate stronger learning ability \cite{szegedy2015going, ding2021repvgg} despite being more computationally expensive. 
Re-parameterization is a significant technique that has been frequently used for better performance while reducing computational costs in the inference stage. 
Some successful implementations like RepLKNet \cite{ding2022scaling}, RepVGG \cite{ding2021repvgg} and ACNet \cite{ding2019acnet} show that re-parameterization can be adopted to integrate the multiple branches equivalently and improve inference efficiency. 

For our proposed RKF model, a multi-branch structure is applied within each convolution layer. By virtue of the re-parameterization approach, the model can be equally slimmed to the original single-branch structure, thus reducing the computation cost considerably.

\subsection{Knowledge Distillation (KD)}
Knowledge Distillation (KD) aims to supervise a light student model with a cumbersome teacher model and can be adopted for model compression.
% The major concern of KD is how to transfer the knowledge from the large model to the small model. Based on the forms of knowledge, current methods can be summarized into three categories: response-based, feature-based, and relation-based \cite{gou2021knowledge}.
KD is considered to be an effective method in multiple tasks, including classification \cite{hinton2015distilling,li2017learning, peng2019few}, object detection \cite{chen2017learning, li2017mimicking} and semantic segmentation \cite{he2019knowledge, dou2020unpaired}.
In this paper, we propose to leverage the distillation strategy for auxiliary knowledge when training our proposed RKF model.

\section{Methods}
\label{methods}
In this section, we describe each component of our method, i.e. Rotated Kernel Fusion \textbf{(RKF)} and Multi-oriented Feature Aggregation \textbf{(MOFA)}.
% First Rotated Kernel Fusion \textbf{(RKF)} is proposed 
% which imposes rotations on each convolution kernel and will not introduce any extra computation cost in the inference phase due to the subsequent re-parameterization process.
% Then another structure, Multi-oriented Feature Aggregation \textbf{(MOFA)}, is presented which integrates features extracted from multiple rotated versions of the input images.
We make the most of MOFA by applying it as a teacher to supervise the RKF model in a manner of knowledge distillation. 

\begin{figure}
    \centering
    \includegraphics[width=0.48\textwidth]{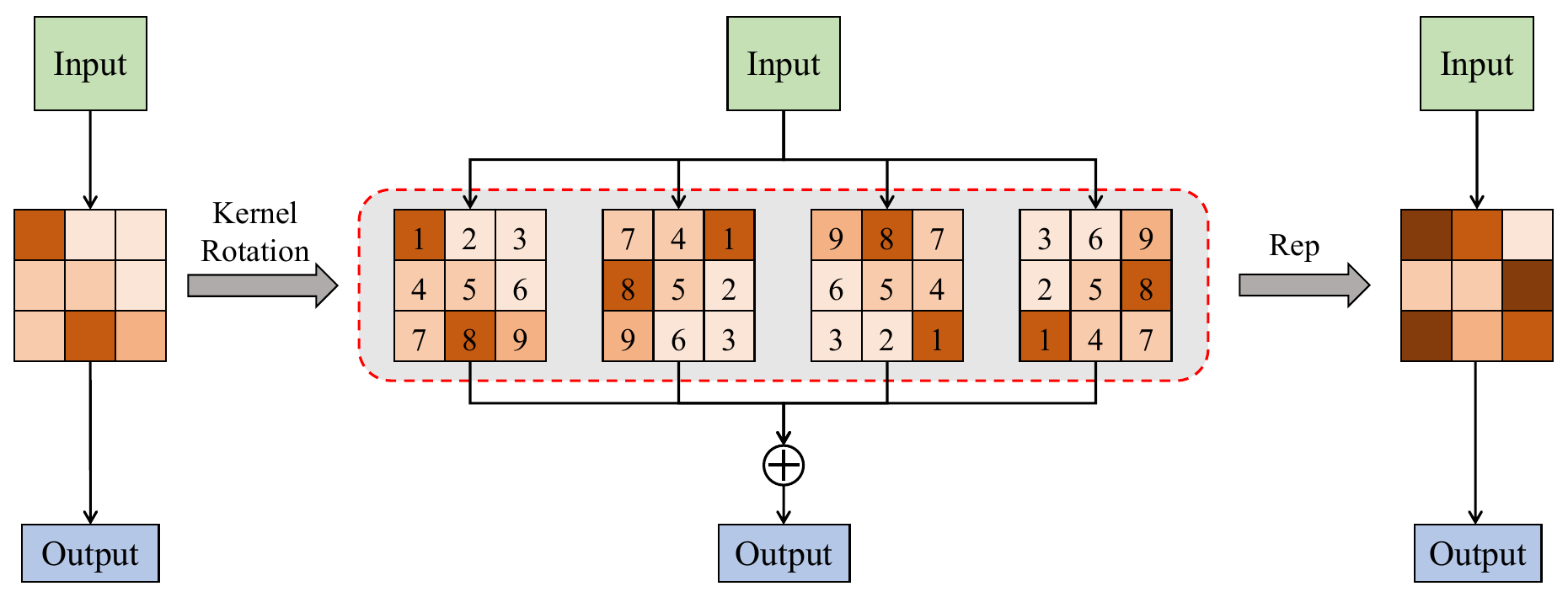}
    
    \caption{The structure of RKF convolution in the case of $3\times3$ kernel. RKF imposes kernel rotation on the original kernel, and fuses features extracted by multi-oriented kernels. Re-parameterization can be further leveraged to convert the multiple kernels to one single kernel. 
    }
    \vspace{-8pt}
    \label{RKF}
\end{figure}

\begin{figure}[t]
    \centering
    \includegraphics[width=0.3\textwidth]{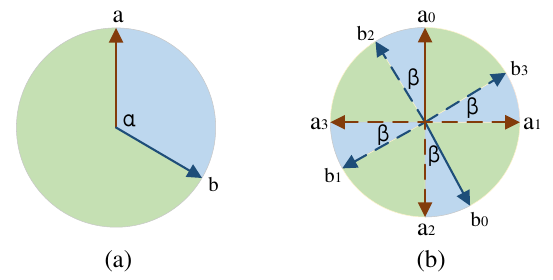}
    \vspace{-10pt}
    \caption{Visualized explanation of MOFA in the case of four rotation transformations on the input pair of images. (a) The two original corresponding keypoints $a$ and $b$ with an orientation gap of $\alpha$. (b) The transformed corresponding keypoints on two images under four rotation transformations. The orientations of $a$ and $b$ are transformed to $a_0$, $a_1$, $a_2$, $a_3$ and $b_0$, $b_1$, $b_2$, $b_3$, respectively, generating multiple correspondence pairs with an orientation gap of $\beta$.
    Instead of matching the two original keypoints with an orientation gap of $\alpha$ on two images, the model only needs to consider several correspondence pairs with a smaller gap of $\beta$ on the transformed keypoints.}
    % Suppose the two corresponding keypoints are $a_0$ and $b_0$. With rotation transformations, the orientation of $a_0$ and $b_0$ are transformed to $a_1$, $a_2$, $a3$ and $b_1$, $b_2$, $b_3$, respectively.
    \vspace{-15pt}
    \label{intuition}
\end{figure}

\subsection{Feature correspondences}
\paragraph{General loss for detection-and-description} 
We follow most existing detection-and-description approaches like D2-Net \cite{dusmanu2019d2} and ASLFeat \cite{luo2020aslfeat},
and adopt the loss function as Eq.\ref{eq:d2} to jointly optimize the detection and description objectives:
\begin{equation}
    \mathcal{L}_{ori} = \frac{1}{\mathcal{C}}\sum_{c\in\mathcal{C}}\frac{s_cs'_c}{\sum_{q\in\mathcal{C}}s_qs'_q}\mathcal{M}(\bm{f}_c, \label{eq:d2} \bm{f}'_c),
\end{equation}
where $\mathcal{C}$ is the ground-truth correspondence set, $s_c$ and $s'_c$ are corresponding scores, $\bm{f}_c$ and $\bm{f}'_c$ refer to corresponding descriptors, and $\mathcal{M}(\cdot, \cdot)$ is the ranking loss for representation learning, which can be defined as triplet \cite{2015FaceNet} or contrastive loss \cite{2006Dimensionality}.

\paragraph{Homography Augmentation}
Similar to \cite{parihar2021rord}, we incorporate in-plane rotations in augmentations on the training data to simulate realistic viewpoint changes. 
Given an image pair, we augment either image of the pair using a random homography transformation consisting of 0$-2\pi$ rotations.
This increases the data diversity and enables the model to be exposed to extreme rotations in the training stage.
Besides the in-plane rotations, we also incorporate scaling, skewness and perspectivity transformations in the augmentations as done in \cite{detone2018superpoint}.

\subsection{Rotated Kernel Fusion (RKF)}
\paragraph{Equivariance properties of CNNs}

Let $\mathbf{M_1} \in \mathbb{R}^{
N \times C_1 \times H_1 \times W_1}$, $\mathbf{M_2} \in \mathbb{R}^{
N \times C_2 \times {H_2} \times {W_2}}$ be the input and output feature map, respectively.
We use $*$ as the convolution operator, and $\mathbf{W}\in \mathbb{R}^{
C_2 \times C_1 \times k \times k}$ 
to denote the kernel
of a $k \times k$ convolution kernel with $C_1$ input channels and $C_2$ output channels.  
Therefore the regular convolution can be denoted as $\mathbf{M_2}=\mathbf{M_1}*\mathbf{W}$.
% Given the input image or feature map $\mathbf{f}\in \mathbb{R}^{
% C\times H\times W}$, convolution kernel $\mathbf{\Phi}\in \mathbb{R}^{
% k \times k}$, the regular convolution can be denoted as $\mathbf{f}*\mathbf{\Phi}$.
Convolution layers can be used effectively in a deep network because all the layers are translation equivariant: shifting the input and then feeding it through the convolution layer is the same as feeding the original input through the same layer and then shifting the result, which can be represented as follows:
\begin{equation}
    [L_t\mathbf{M_1}]*\mathbf{W} = L_t[\mathbf{M_1}*\mathbf{W}],
\end{equation}
where $L_t$ denotes the translation transformation.

However, CNNs are not rotation invariant inherently, which means rotating the input and then convolving it with a filter is not the same as first convolving and then rotating the result.
In fact, the convolution of a rotated input $L_r\mathbf{M_1}$ with a filter $\mathbf{W}$ is equal to the rotation of the result obtained by original input $\mathbf{M_1}$ convolving with the inverse-rotated filter $L_r^{-1}\mathbf{W}$ \cite{cohen2016group}.
This can be formulated as:
\begin{equation}
    [L_r\mathbf{M_1}]*\mathbf{W} = L_r[\mathbf{M_1}*L_r^{-1}\mathbf{W}],
    \label{eq:rotation}
\end{equation}
where $L_r$ denotes the rotation transformation.

\paragraph{Motivation and Design}
Because CNNs are not rotation invariant inherently, they are susceptible to large rotation changes.
To improve the inherent nature of convolution layers, inspired by Group equivariant CNN (G-CNN) \cite{cohen2016group}, we propose Rotated Kernel Fusion (RKF) which imposes rotations on each convolution kernel and fuses the feature maps extracted by multi-oriented kernels, shown in Fig. \ref{RKF}. The major difference from \cite{cohen2016group} is that we choose summation instead of concatenation for the fusion of the multi-oriented feature maps. Using summation operation, multiple kernels can be merged into one single kernel with the original shape by the subsequent re-parameterization operation to obtain higher efficiency and convenience in deployment.

We split the rotation range into N-1 intervals and denote the rotation transformation on each kernel $\mathbf{W}$ as $L_r^{\theta_n}$ where $\theta_n=\frac{2\pi}{N}n, 0\leq n \leq N-1$ refers to the angle of rotation. 
\begin{equation}
    L_r^{\theta_n} = \begin{bmatrix}
       \cos(\theta_n) & -\sin(\theta_n)  \\
       \sin(\theta_n) & \cos(\theta_n) 
       \end{bmatrix}
\end{equation}
This transformation matrix acts on points in $\mathbb{Z}^2$ (pixel coordinates) by multiplying the matrix $L_r^{\theta_n}$ by the coordinate vector $x(u, v)$ of a point:
\begin{equation}
    L_r^{\theta_n}x = \begin{bmatrix}
       \cos(\theta_n) & -\sin(\theta_n)  \\
       \sin(\theta_n) & \cos(\theta_n) 
       \end{bmatrix}            
    \begin{bmatrix}
u \\ v  \end{bmatrix}
\end{equation}

The RKF convolution is denoted as:
\begin{equation}
    % \hat{\mathbf{X}}_{RKF} = z  \mathbf{f} * \Phi + (1-z) \sum_{i=0}^{n-1}\left(\mathbf{f} * (L_r^{\theta}\Phi)\right),
    \mathbf{M_2}=\mathbf{M_1}*\mathbf{W}_{RKF} =  \sum_{n=0}^{N-1}\left(\mathbf{M_1} * [L_r^{\theta_n}\mathbf{W}]\right), \theta_n = \frac{2\pi}{N}n.
    \label{eq:rkf}
\end{equation}
To validate the equivariance of RKF, we suppose the RKF convolution layer is fed with a rotated image or feature map $L_r^{\theta_m}\mathbf{M_1}$, $0\leq m \leq N-1$, then the result is:
\begin{equation}
\begin{aligned}
[L_r^{\theta_m}\mathbf{M_1}]*\mathbf{W}_{RKF}
&=\sum_{n=0}^{N-1}\left([L_r^{\theta_m}\mathbf{M_1}] * [L_r^{\theta_n}\mathbf{W}]\right).
    \label{eq:rotate_image}
\end{aligned}
\end{equation}
Based on Eq. \ref{eq:rotation}, we let $\mathbf{W}=L_r\mathbf{W}$ and obtain:
\begin{equation}
    [L_r\mathbf{M_1}]*[L_r\mathbf{W}] = L_r[\mathbf{M_1}*\mathbf{W}].
    \label{eq:rotation2}
\end{equation}

Then based on Eq. \ref{eq:rotation2} and the property that rotation by $\theta_{-m}$ is the same as the rotation by $\theta_{N-m}$,
we can rewrite Eq. \ref{eq:rotate_image} as:
\begin{equation}
\begin{aligned}
[L_r^{\theta_m}\mathbf{M_1}]*\mathbf{W}_{RKF}
    &=L_r^{\theta_m}\sum_{n=0}^{N-1}\left(\mathbf{M_1} * [L_r^{\theta_n-m}\mathbf{W}]\right)\\
     &=L_r^{\theta_m}\sum_{n=-m}^{N-1-m}\left(\mathbf{M_1} * [L_r^{\theta_n}\mathbf{W}]\right)\\
     &=L_r^{\theta_m}\sum_{n=0}^{N-1}\left(\mathbf{M_1} * [L_r^{\theta_n}\mathbf{W}]\right)\\
     &=L_r^{\theta_m}[\mathbf{M_1}*\mathbf{W}_{RKF}]. 
    \label{eq:rotate_image2}
\end{aligned}
\end{equation}
Therefore, RKF enables the convolution to be equivariant to rotation, which means rotating the input and then feeding it through the layers is the same as feeding the original input and then rotating the result. 
Accordingly, the local feature descriptors can obtain better invariance to rotation: the local descriptors are the same for two corresponding points on a pair of images with relative rotation.
Although the above conclusions are based on discrete rotation angles, 
in fact,  
for an image rotated by a randomly sampled angle, the orientation difference between the rotated image and the nearest rotated filter is limited to at most $\frac{\pi}{N}$, enhancing the robustness of rotations.

We call the ordinary model with regular convolution layers \textbf{\textit{base model}}, and the new model with RKF convolution layers \textbf{\textit{RKF model}} in the following sections.

\paragraph{Re-parameterization}
Because those operations introduced by RKF are linear within each convolution layer, we can further use re-parameterization to fuse multiple trained kernels into a single one for inference. In this way, RKF will not alter the original CNN architecture and introduce extra computation in the inference phase. According to Eq. \ref{eq:rkf}, we can obtain the final single kernel $\mathbf{W}_{RKF} \in \mathbb{R}^{
C_2 \times C_1 \times k \times k}$ as follows:
\begin{equation}
    \mathbf{W}_{RKF} =  \sum_{n=0}^{N-1}\left( L_r^{\theta_n}\mathbf{W}\right)
\end{equation}

% Overall, instead of using the base model as the student model, we adopt the enhanced student model, which is the RKF model, as the student to learn from MOFA, and we refer to the distilled model as Distilled Rotated Kernel Fusion (DRKF).

\subsection{Distilled Rotated Kernel Fusion (DRKF)}

\paragraph{Multi-oriented Feature Aggregation (MOFA)}
While RKF leads to significant performance gains in local feature matching under extreme rotation variations, we observe that 
the training optimization is faced with several drawbacks:
1) Different rotated kernels share the same parameters of the original filter, therefore imbalanced training might be caused when the contribution of some rotated kernels overweight others;
2) For the purpose of re-parameterization, no non-linear operations are involved before the integration of the results from different rotated kernels.
To overcome these limitations, we propose another structure called Multi-oriented Feature Aggregation (\textbf{MOFA}) to provide auxiliary supervision for RKF. MOFA conducts post-processing on the base model trained with homography augmentations.
% thus involving no additional training process.
Formally, we use $F_{base}(\cdot)$ to represent the
base model that projects the input to the descriptor or detector score
map and use $F_{MOFA}(\cdot)$ to denote the projection of MOFA. The input image $\bm{I}$ is rotated by $N$ different angles respectively, and then fed into the trained base model to get the corresponding outputs, which are further aggregated as follows:
\begin{equation}
    F_{mofa}(\bm{I}) = \frac{1}{N}\sum_{n=0}^{N-1}{L_r^{-\theta_n}}\left(F_{base}(L_r^{\theta_n}(\bm{I}))\right),  \theta_n = \frac{2\pi}{N}n.
    \label{eq:mofa}
\end{equation} 
% To be specific, input images are rotated with multiple orientations, fed into the trained base model, 
% and the features based on multi-oriented inputs are subsequently integrated.
Therefore each rotated branch of MOFA incorporates non-linearity and contributes independently and equally, which compensates for the limitations of RKF training.
We can understand MOFA intuitively in Fig. \ref{intuition}: multiple rotation transformations convert the original large orientation gap of two corresponding points on the input pair of images to a smaller gap, which generates features that are more robust to large rotation variations.

% the output of base model can be formulated as follows, 
% \begin{equation}
%     F_{base}(\bm{I}) = f(\bm{I}).
%     \label{eq:base}
% \end{equation}
% MOFA can be theoretically formulated as:
% \begin{equation}
%     \mathcal{F}_{mofa}(\bm{I}) = \frac{1}{2\pi}\int_{0}^{2\pi}g^{-\theta}\left(F_{base}(g^{\theta}(\bm{I}))\right)d\theta, \label{eq:integral}
% \end{equation}
% where $g^{\theta}$ refers to rotation operation with radian $\theta$, $g^{-\theta}$ is the corresponding inverse operation, and $F_{base}(\cdot)$ represents the base model that projects the input to the descriptor or score map. According to Eq. \ref{eq:integral}, features extracted from all-oriented inputs are aggregated, therefore the rotation robustness can be improved.

% However, integration in Eq. \ref{eq:integral} is intractable, and an alternative pipeline is shown in Fig. \ref{pipeline}(a).

\paragraph{Knowledege Distillation}
For MOFA, the increase in rotation invariance comes at a cost of multiple feedforwards, making it computationally expensive for embedded devices.
On the other hand, the training optimization of computationally efficient RKF is affected by the tangled branches of rotated kernels and the lack of non-linear activation before feature fusion.
Therefore, we propose to utilize a knowledge distillation strategy and take advantage of MOFA to provide auxiliary supervision to the RKF model. 
Suppose the normalized dense descriptor map and detector score map generated by the MOFA is $D^{(t)} \in {\frac{H}{r}}\times{\frac{W}{r}}\times{C}$ and $S^{(t)} \in {H}\times{W}\times{1}$ respectively, and those generated by the RKF are denoted as $D^{(s)} \in {\frac{H}{r}}\times{\frac{W}{r}}\times{C}$ and $S^{(s)} \in {H}\times{W}\times{1}$ respectively, where $H$ and $W$ refer to height and width of the input image respectively, and $r$ is the downsampling rate. The distillation loss of descriptors is defined in the form of mean distance error, which is formulated as:
\begin{equation}
    \mathcal{L}_{dis}^{(desc)} = \frac{r^2}{HW}\sum_{i,j}\left\|D_{ij}^{(s)} - D_{ij}^{(t)}\right\|_2. \label{eq:desc}
\end{equation}
And the distillation loss of the score map is in the form of local cross-entropy loss. We reshape the two score maps $S^{(s)}$ and $S^{(t)}$ to ${\frac{H}{r}}\times{\frac{W}{r}}\times{r^2}$ with inverse pixel shuffle \cite{2016Real} operation, and define the distillation loss of the score map as:
\begin{equation}
    \mathcal{L}_{dis}^{(score)} = -\frac{r^2}{HW}\sum_{i,j}\sum_{k=1}^{r^2}P_{ijk}^{(t)}\log{P_{ijk}^{(s)}}, \label{eq:score}
\end{equation}
where
\begin{equation}
P_{ijk} = \exp(S_{ijk}) / \sum_{k}\exp(S_{ijk}).
\end{equation}
The total distillation loss is a combination of Eq. \ref{eq:desc} and Eq. \ref{eq:score}, which is
\begin{equation}
    \mathcal{L}_{dis} = \mathcal{L}_{dis}^{(desc)} + \lambda_1{\mathcal{L}_{dis}^{(score)}}.
    \label{eq:dis}
\end{equation}

Therefore, the total training loss of our method is formulated as:
\begin{equation}
    \mathcal{L} = \mathcal{L}_{ori} + \lambda_2{\mathcal{L}_{dis}}.\label{eq:total}
\end{equation}

In this way, we obtain the distilled
RKF model, which is referred to as \textbf{\textit{DRKF model}}. Similarly, we can also obtain the distilled base model by transferring knowledge from MOFA to the base model, which is called \textbf{\textit{DBase model}} in our following sections.

\section{Experiments}
\subsection{Datasets}
\paragraph{GL3D \cite{shen2018mirror}}
This dataset is only used for training. It contains 181,280 high-resolution images in 1073 different scenes, along with the image matches, the intrinsics, and extrinsics, as well as the depth data. 

\paragraph{HPatches \cite{balntas2017hpatches}}
Out of the 116 available sequences, we selected 108 as implemented in \cite{dusmanu2019d2} for evaluation. Each sequence consists of 6 images of illumination changes (52 sequences) or viewpoint changes (56 sequences).
The original dataset is called Standard HPatches in our experiments. 
Besides, we also use the Rotated HPatches dataset constructed by \cite{parihar2021rord} which is the augmented version of the original dataset with random in-plane rotation of images from 0 to $2\pi$.

\paragraph{DiverseBEV}

\begin{figure} 
\centering    
\includegraphics[width=0.25\textwidth]{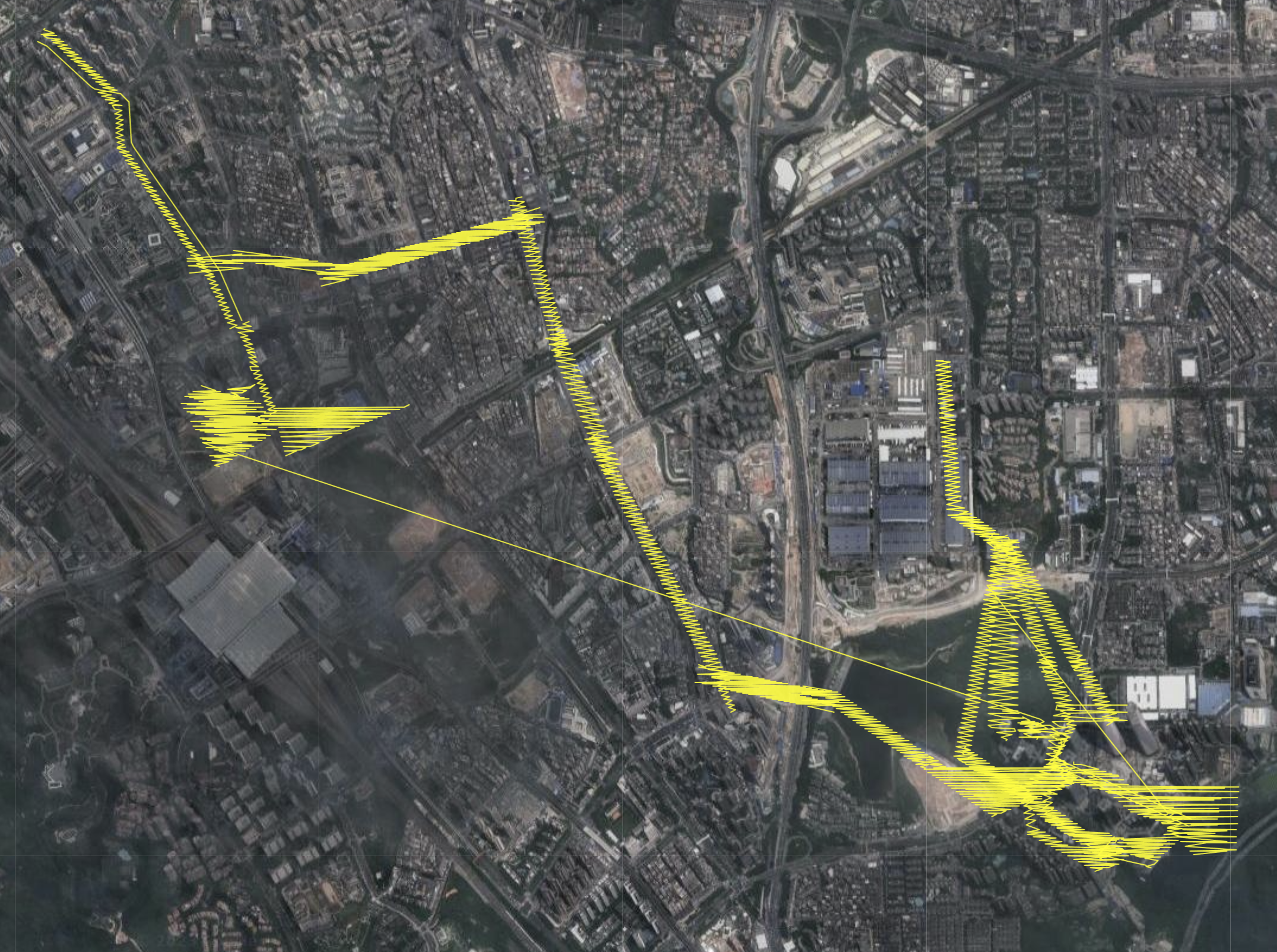}  
\vspace{-10pt}
\caption{The flight trajectory of the DiverseBEV dataset.}  
\vspace{-15pt}
\label{bev} 

\end{figure}
Currently, most existing works for local feature matching are based on ground-view datasets \cite{balntas2017hpatches, parihar2021rord, dusmanu2019d2, luo2020aslfeat}.
Compared to the ground-view cameras, the aerial-view platforms, e.g., drones, capture the scene image from the sky with a wider angle, less occlusion and more comprehensive viewpoints, therefore the aerial-view data can be useful to investigate the rotation invariance of local descriptors.
Due to the lack of publicly available aerial-view datasets with high viewpoint changes and camera rotations, we introduce a new dataset called DiverseBEV. This dataset comprises bird's eye view outdoor scenarios captured during the drone's fight, thus exhibiting large or even extreme rotation variations. First, 5292 images at a resolution of $5472\times 3648$ are collected using an FC6310R camera on DJI M300 aircraft with an average altitude of 207m in Shenzhen, China.
% These images are captured during the drone's fight, thus exhibiting extreme rotation variations.
We take RTK as a priori, implement 3D reconstruction with COLMAP \cite{schonberger2016structure,schonberger2016pixelwise} with GPS Position Prior \footnote{https://github.com/Vincentqyw/colmap-gps}, and obtain $K$, $R$, $T$ and depth for each image which is taken as ground truth intrinsic, extrinsic parameters, and depth information.
We traverse all images and select all image pairs with a distance from 20m to 50m according to the $T$ of each image. Then, based on the rotation angle $R$ of each image, the relative rotation angle of each pair of images is obtained. For relative rotation angles ranging from $0^{\circ}$ to $360^{\circ}$, around 25 image pairs are randomly sampled every $10^{\circ}$ for evaluation, thus 901 image pairs are sampled in total which constitute our DiverseBEV dataset.
For this dataset, evaluation is done using the model trained only on GL3D, therefore demonstrating the generalizability of our proposed method.
For a depiction of the flight path see Fig. \ref{bev}.

% To analyze the effectiveness of rotation augmentation in the training phase,
% we implement the training process under the following two settings:
% \begin{itemize}
%     \item Train \textit{w/o rot}: Train without rotation augmentations.
%     \item Train \textit{w rot}: Train with rotation augmentations of radians in $[0, 2\pi)$;
% \end{itemize}
\subsection{Experimental Settings}
In the training phase, we first train the base model with homography
augmentations for 400,000 iterations with a learning rate of 0.1. The trained base model is used to construct MOFA. Then the RKF model is trained with 100,000 iterations with the supervision of MOFA.
We set $n=4$ in Eq. \ref{eq:rkf} and Eq. \ref{eq:mofa}, therefore the rotation angles in MOFA and RKF are $0, \pi/2, \pi, 3\pi/2$.
 For loss hyperparameters, we set $\lambda_1=\lambda_2=1$ in Eq. \ref{eq:dis} and Eq. \ref{eq:total}.

\subsection{Evaluation Protocol}
\paragraph{Mean Matching Accuracy (MMA)} 
We follow the standard protocol \cite{dusmanu2019d2} for evaluating nearest neighbor feature matching on HPatches dataset. 
For each image pair of HPatches, we match the descriptors extracted by each method using nearest neighbor search with the mutual test. 
A match is considered to be correct if the back-projection of the keypoint 
 estimated by the ground truth homography matrix falls within a given pixel threshold.
We vary the threshold and record the mean matching accuracy (MMA) \cite{mikolajczyk2005performance} over all pairs, which is the average percentage of correct matches per image pair \footnote{Evaluation code: https://github.com/mihaidusmanu/d2-net}.

\paragraph{Mean Average Accuracy (MAA)}
For DiverseBEV, we follow \cite{jin2021image} and we compute the angular difference, in degrees, between the estimated and
ground-truth quaternion vectors of two
cameras. Distance difference is computed between the estimated and
ground-truth translation vectors. We then threshold the differences over varying values for all possible pairs of images, which renders a curve. We compute the
Mean Average Accuracy (MAA) by integrating this curve
up to a maximum threshold ($10^{\circ}$ and 10m for the angular and distance error respectively) \footnote{Evaluation code: https://www.kaggle.com/competitions/image-matching-challenge-2022}.

\subsection{Comparison with SOTA approaches}
In this part, we first compare our proposed method with SIFT \cite{lowe2004distinctive} and existing state-of-the-art learning-based approaches. It is notable that DRKF after re-parameterization shares the same results as DRKF in TABLE \ref{hpatches} and \ref{bev_table}. 
% Here, we mainly focus on feature extraction stage, therefore we do not compare our approach to SOTA feature matching approaches, including SuperGlue \cite{sarlin2020superglue}, LofTR \cite{sun2021loftr}, CoTR \cite{jiang2021cotr}, ClusterGNN \cite{shi2022clustergnn}, etc. 

% Besides, for SuperPoint, data augmentation is essential in training for pseudo label generation, therefore we do not evaluate the model trained without rotation augmentations.

\paragraph{HPatches}
\begin{table}[]
\small
\setlength\tabcolsep{3.5pt}
    \centering
    \caption{Evaluation Results for MMA on the HPatches dataset using pixel thresholds 6/8/10.}
    \vspace{-10pt}
    \begin{threeparttable}
        \resizebox{0.45\textwidth}{!}{
        \begin{tabular}{ccccccccc}
        \toprule
        Methods & Rotation Aug &  Standard & Rotated & Average \\
        \midrule
        SIFT \cite{lowe2004distinctive} & \textbf{--} & 0.52/0.54/0.54 & 0.51/0.51/0.52 &  0.54/0.53/0.53  \\
        SuperPoint \cite{detone2018superpoint} & \CheckmarkBold & 0.69/0.71/0.73 & 0.21/0.22/0.22 & 0.45/0.46/0.48 \\
        D2-Net \cite{dusmanu2019d2} & \XSolidBrush& 0.73/0.81/0.84 & 0.17/0.20/0.22 & 0.45/0.50/0.53\\
        
        ASLFeat \cite{luo2020aslfeat} & \XSolidBrush& 0.81/0.82/0.82 & 0.22/0.24/0.24 & 0.52/0.53/0.53\\
        GIFT \cite{liu2019gift} & \CheckmarkBold & 0.83/0.86/0.87 & 0.43/0.44/0.44 & 0.63/0.65/0.66 \\
        RoRD \cite{parihar2021rord} & \CheckmarkBold  & 0.79/0.84/0.86 & 0.48/0.59/0.64 & 0.64/0.72/0.75 \\
        
        \midrule
        (Ours) DRKF* &\CheckmarkBold &  \textbf{0.84/0.86/0.87} & \textbf{0.80/0.81/0.82} & \textbf{0.82/0.84/0.85}  \\
        
        \bottomrule
        \end{tabular}}
        \begin{tablenotes} %添加此处
		\item * Same results for DRKF after re-parameterization.
		\end{tablenotes} %添加此处
    \end{threeparttable}
    \vspace{-5pt}
    \label{hpatches}
    % \vspace{-5pt}
\end{table}

Different methods are compared on Standard HPatches, Rotated HPatches, and the average of two in TABLE \ref{hpatches}, in terms of
 the mean matching accuracy (MMA) of all pairs over pixel thresholds 6, 8, and 10. This setting for thresholds is the same as implemented in \cite{parihar2021rord}.
 The "Rotation Aug" column denotes whether the rotation augmentation is adopted during the training process.
 For SIFT \cite{lowe2004distinctive}, SuperPoint \cite{detone2018superpoint}, D2-Net \cite{dusmanu2019d2} and RoRD \cite{parihar2021rord}, we directly quote the statistics reported in \cite{parihar2021rord}.
 To validate the performance of ASLFeat \cite{luo2020aslfeat} and GIFT \cite{liu2019gift} we use the trained model provided by the original authors.
 In this part, we mainly focus on feature
extraction, therefore we do not compare our approach
to SOTA feature matching approaches, e.g. SuperGlue \cite{sarlin2020superglue} and LoFTR \cite{sun2021loftr}.
According to the results, it can be observed that the performance of SuperPoint \cite{detone2018superpoint}, D2-Net \cite{dusmanu2019d2} and ASLFeat \cite{luo2020aslfeat} degrades greatly on rotation-augmented datasets, while SIFT shows steady performance under random rotation due to its elaborate design for orientation prediction. 
GIFT \cite{liu2019gift} and RoRD \cite{parihar2021rord} are trained with homography
augmentations and are specifically designed to improve rotation invariance, therefore they overtake other methods on Rotated Hpatches.
% uses group features, which are extracted on the transformed images, and adopts group CNNs to encode such features to construct transformation-invariant descriptors.
% RoRD \cite{parihar2021rord} learns invariant descriptors through data augmentation and orthographic viewpoint projection, and obtains better generalization when a significant level of rotation is introduced to the test data. 
However, our DRKF still surpasses them by a large margin, for example, 32\% better than RoRD with regards to MMA over a threshold of 6 on Rotated Hpatches.
It is notable that the proposed DRKF even outperforms SIFT considerably when testing on the rotation-augmented dataset while retaining the performance on standard datasets. 
This demonstrates that DRKF effectively enables the network to extract more rotation-invariant features.

\paragraph{DiverseBEV}
\begin{table}[]
\small
\setlength\tabcolsep{3.5pt}
    \centering
    \caption{Evaluation Results for MAA on the DiverseBEV dataset. The numbers are obtained by integrating the corresponding curves in Fig. 
    \ref{bev_compare}. }
    \vspace{-10pt}
    \begin{threeparttable}
        \resizebox{0.45\textwidth}{!}{
        \begin{tabular}{ccccc}
        \toprule
        Methods & Rotation Aug &  \makecell{MAA \\($10^{\circ}\&10$m)} & \makecell{MAA\\($10^{\circ}$)} & \makecell{MAA\\(10m)} \\
        \midrule
        SIFT \cite{lowe2004distinctive} & \textbf{--} & 0.3096 & 0.5741 & 0.3103  \\
        SuperPoint \cite{detone2018superpoint}+SuperGlue \cite{sarlin2020superglue} & \CheckmarkBold & 0.1388 & 0.2574 & 0.1428 \\
        D2-Net \cite{dusmanu2019d2} & \XSolidBrush& 0.0319 & 0.0807 & 0.0331\\
        
        ASLFeat \cite{luo2020aslfeat} & \XSolidBrush& 0.0789 & 0.1548 & 0.0809\\
        LoFTR \cite{sun2021loftr} & \XSolidBrush& 0.0583 & 0.1226 & 0.0592\\
        GIFT \cite{liu2019gift} & \CheckmarkBold & 0.1200 & 0.2410 & 0.1216 \\
        RoRD \cite{parihar2021rord}& \CheckmarkBold  & 0.1833 & 0.5110 & 0.1848 \\
        
        \midrule
        (Ours) DRKF* &\CheckmarkBold &  \textbf{0.3914} & \textbf{0.7523} & \textbf{0.3966} \\
        
        \bottomrule
        \end{tabular}}
        \begin{tablenotes} %添加此处
		\item * Same results for DRKF after re-parameterization.
		\end{tablenotes} 
    \end{threeparttable}
    \label{bev_table}
    % \vspace{-15pt}
\end{table}

\begin{figure*} 
\centering    
\includegraphics[width=0.59\textwidth]{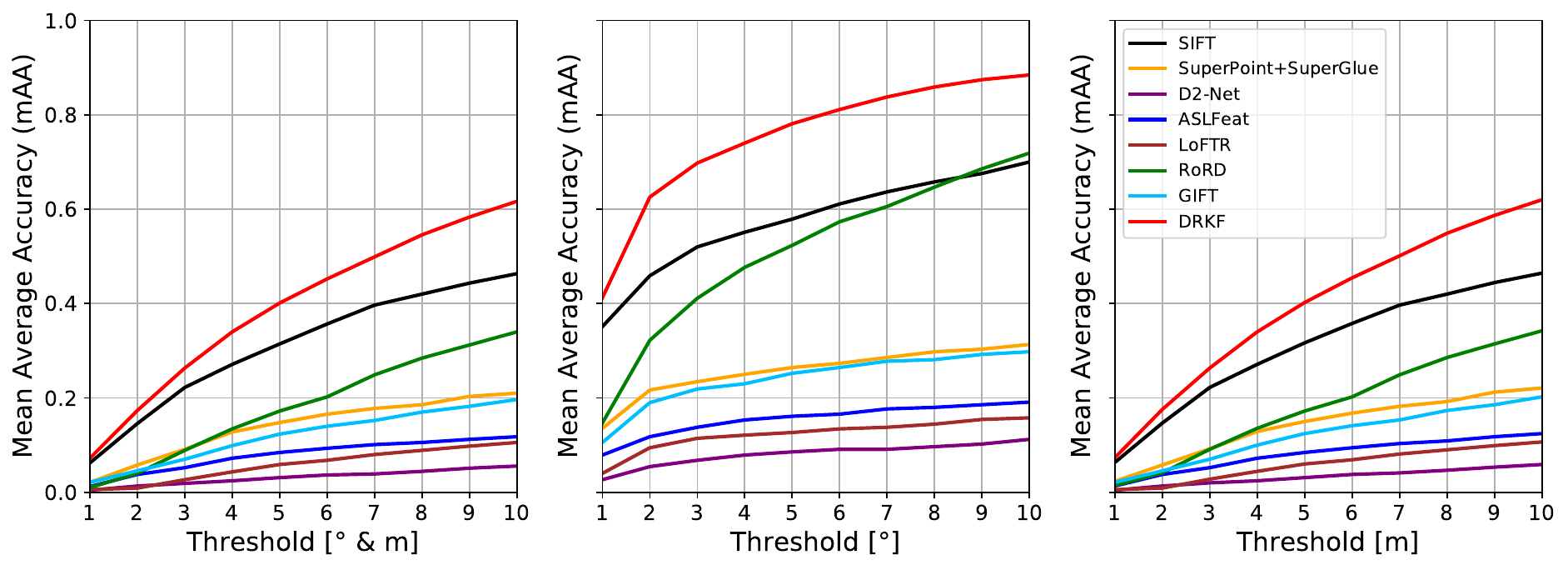}     
\vspace{-10pt}
\caption{Mean average accuracy (MAA) curves on the DiverseBEV dataset. The left shows the MAA of image pairs satisfying both angular and distance thresholds. 
The middle and right are the MAA as a function of the angular threshold and distance threshold, respectively. Our DRKF outperforms other SOTA methods on the DiverseBEV dataset.}   
\label{bev_compare}  
\vspace{-5pt}
\end{figure*}

We resize images of the DiverseBEV dataset to $600 \times 1000$ and compare DRKF to other SOTA methods for Mean Average Accuracy (MAA) curves on the DiverseBEV dataset, shown in Fig. \ref{bev_compare}. For SIFT, we use its OpenCV implementation. For learning-based methods, we use the trained model provided by the original authors. In this part, we also involve state-of-the-art feature matching methods like SuperGlue \cite{sarlin2020superglue} and LoFTR \cite{sun2021loftr}.
It can be seen that our DRKF obtains the largest MAA over varying angular or distance thresholds.
TABLE \ref{bev_table} shows the integration result of these curves
up to $10^{\circ}$ or/and 10m. The third column shows the integration of MAA over both angular and distance thresholds. The last two columns are with regard to angular threshold up to {$10^\circ$} and distance threshold up to 10m, respectively. It can be seen from the third column that SIFT overtakes all previous learning-based methods, while our method obtains 8.18\% performance gain over SIFT on average MAA.
Since the DiverseBEV dataset displays large rotation variations of the camera during the drone's flight, the experiments can show that although our method is based on in-plane rotation operations, it also works for real-world out-of-plane rotations and can perform stably under extreme viewpoint changes.

\begin{table}[]

\small
    \centering
    \small
    \caption{Inference Time on Nvidia TX2 }
    % \vspace{-5pt}
    \begin{threeparttable}
    \resizebox{0.35\textwidth}{!}{
        \begin{tabular}{cc}
        \toprule
         Methods &  Inference Time (ms)\\
        \midrule
        SuperPoint 
 \cite{detone2018superpoint} + SuperGlue \cite{sarlin2020superglue} & 110\\
        RoRD \cite{parihar2021rord} & 229 \\
        ASLFeat \cite{luo2020aslfeat} & 56 \\
        \midrule
        MOFA & 207 \\
        Base & 58 \\
        DRKF & 213 \\
        DRKF (*Rep) & 57 \\
        \bottomrule
        \end{tabular}}
        \begin{tablenotes} %添加此处
		\item *Rep refers to re-parameterization.
		\end{tablenotes} %添加此处
    \end{threeparttable}
    \label{efficiency}
    \vspace{-15pt}
\end{table}

\begin{figure*}[h!] 
\centering    
\includegraphics[width=0.68\textwidth]{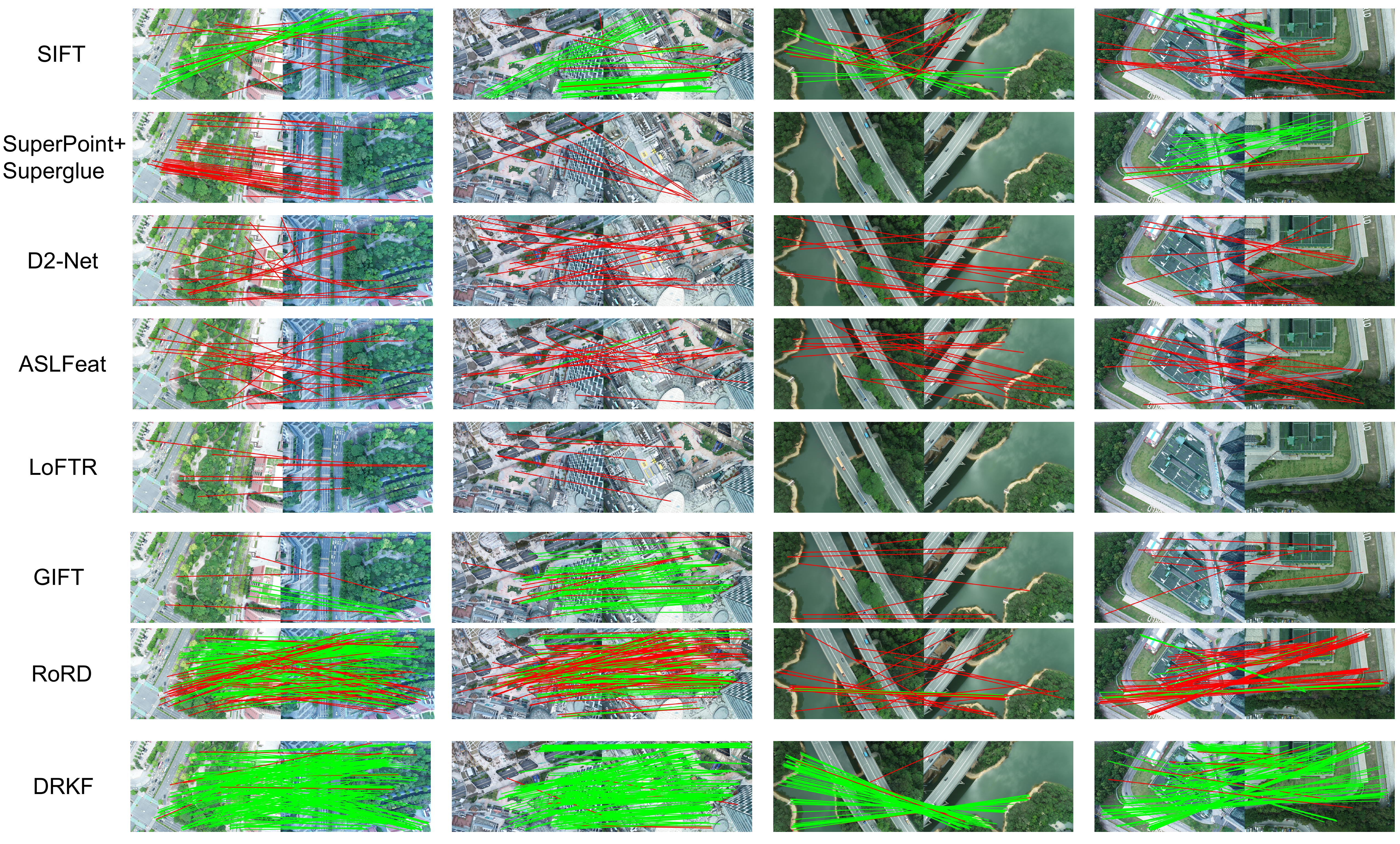} 
\vspace{-10pt}
\caption{Qualitative results on the DiverseBEV Dataset. Our method DRKF (bottom) outperforms all other methods by a significant margin. These correspondences are filtered by RANSAC-based geometric verification to obtain a final set of feature correspondences. A match is
considered to be correct if the backprojection of the keypoint
estimated using the ground truth $K$, $R$, $T$ and depth falls
within a pixel threshold of $1.5$. We use red lines for wrong matches, and green for correct matches.}   
\vspace{-10pt}
\label{bev_vis}     
\end{figure*}

\begin{figure} 
\centering    
\subfigure[Standard Hpatches] {
 \label{rotinv-sift}     
\includegraphics[width=0.2\textwidth]{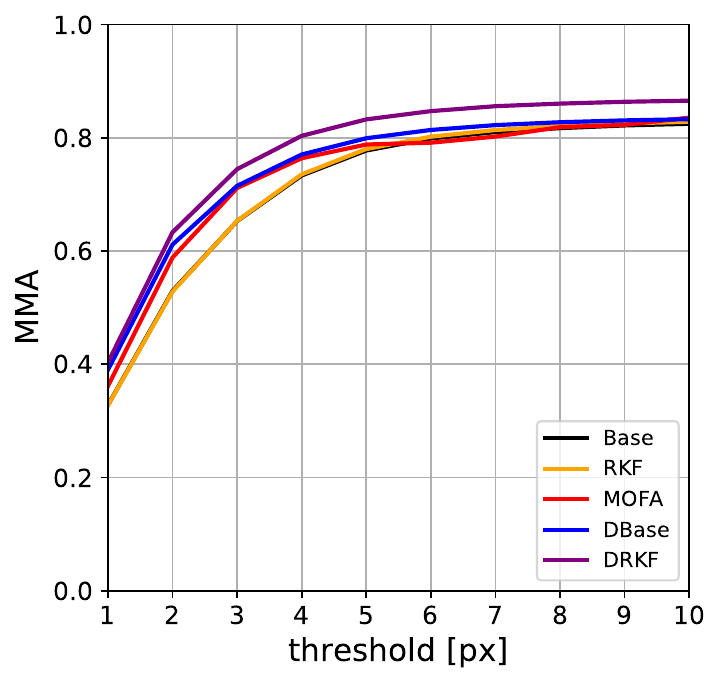}  
}     
\subfigure[Rotated Hpatches] { 
\label{rotinv-lf}     
\includegraphics[width=0.2\textwidth]{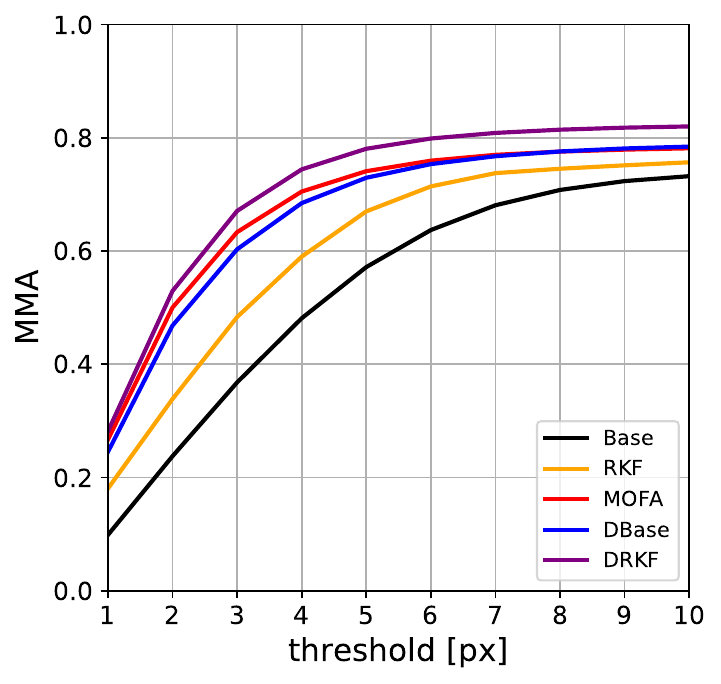}     
}    

%\vspace{-8pt}
\caption{Ablation studies on Standard HPatches and Rotated HPatches image pairs. For each variant, the mean matching accuracy (MMA) as a function of the pixel threshold is shown. DRKF achieves the best overall performance on Rotated Hpatches.}   
\label{ablation}   
\vspace{-18pt}
\end{figure}

\subsection{Ablation Studies}
% Here we validate the effects of MOFA, KD, and RKF respectively through comparing the homography estimation results of the base model in figure 2. (a) trained with and without rotation augmentations, MOFA, distilled based model (DBase), and distilled RKF (DRKF). 
To analyze the effectiveness of each component of our DRKF, we conduct ablation experiments on Standard Hpatches and Rotated Hpatches respectively. There are 5 variants considered in our experiments: the base model, RKF model, MOFA, DBase and DRKF model, which are all introduced in Section \ref{methods}. All experiments are implemented with homography augmentation of training data. 
% \begin{itemize}
%     \item Base: The model with regular convolution layers;
%     \item RKF: The model of which the convolution layers are replaced with rotated kernel fusion (RKF);
%     \item MOFA: Multi-oriented feature aggregation of the base model; 
%     \item DBase: The base model distilled from MOFA;
%     \item DRKF: RKF model distilled from MOFA. 
% \end{itemize} 
The results are shown in Fig. \ref{ablation}, from which it can be concluded that:
\begin{itemize}
    \item For Standard Hpatches, different methods show similar results, while DRKF still achieves the best matching performance under varying thresholds.
    \item The performance of the base model degrades severely on Rotated Hpatches, on the contrary, the matching performance of RKF is more steady. 
    \item MOFA explicitly integrates multi-oriented features of the base model, therefore outperforms the base model on rotation-augmented datasets remarkably.
    \item With the supervision of MOFA, DBase overtakes the base model significantly, while still lagging behind DRKF because of the inherent nature of CNN. DRKF utilizes rotated kernels to enhance the representation ability for rotation invariance, therefore surpasses any other variants, including its teacher, MOFA. 
\end{itemize}
    To summarize, two components of DRKF, i.e. RKF and its teacher MOFA, all contribute to rotation invariance.

\subsection{Qualitative comparison}

Qualitative comparison can provide an intuitive observation of the matching results of different methods. We compare our DRKF with SIFT and learning-based methods, including SuperPoint \cite{detone2018superpoint} + SuperGlue \cite{sarlin2020superglue}, D2-Net \cite{dusmanu2019d2}, ASLFeat \cite{luo2020aslfeat}, LoFTR \cite{sun2021loftr}, GIFT \cite{liu2019gift}, and RoRD \cite{parihar2021rord} in Fig. \ref{bev_vis}. We can see that compared to other methods, DRKF obtains the largest number of correct correspondences under different rotation variations, which shows that it can generalize effectively under large viewpoint changes, and even outperforms SIFT which explicitly assigns orientation to each keypoint.

\subsection{Efficiency Analysis}

For efficiency analysis, we compare the inference time of each method on Nvidia TX2 at the input size of 640$\times$360 in TABLE \ref{efficiency}. It can be seen that MOFA consumes longer inference time than the base model due to muti-feedforward feature aggregation.
By virtue of the reparameterization technique, DRKF can be accelerated, and become computationally comparable to the base model.

% For evaluation, we implement exthe commonly used HPatches dataset \cite{balntas2017hpatches} for homography estimation and YFCC100M dataset \cite{thomee2016yfcc100m} for relative pose estimation. 
% In addition, visualization of correspondences for comparison is offered in Fig. 5.

% \begin{figure*} 
% \centering   
% \subfigure[SIFT correspondences.] {
%  \label{vis_sift}     
% \includegraphics[width=0.31\textwidth]{disp_016_000.jpg}  
% }     
% \subfigure[SuperPoint correspondences (Train: \textit{w} rot).] {
%  \label{vis_sp}     
% \includegraphics[width=0.31\textwidth]{sp_viz.jpeg}  
% }     
% \subfigure[LF-Net correspondences (Train: \textit{w} rot).] { 
% \label{vis_lf}     
% \includegraphics[width=0.31\textwidth]{lf_viz.jpeg}     
% }    
% \subfigure[ASLFeat correspondences (Train: \textit{w} rot).] { 
% \label{vis_asl}     
% \includegraphics[width=0.31\textwidth]{asl_viz.jpeg}     
% }
% \subfigure[DRKF correspondences (Train: \textit{w} rot).] { 
% \label{vis_drkf}     
% \includegraphics[width=0.31\textwidth]{rkf_viz.jpeg}     
% }
% \vspace{1pt}
% \caption{Visualized matching results of different methods under rotation of $\pi$.}     
% \label{visualization} 
% % \vspace{-10pt}
% \end{figure*}

\section{Conclusion}
In this paper, we present an effective framework for creating rotation-invariant descriptors to find correspondences under challenging viewpoint conditions. The key part of our method is RKF, which imposes rotations on each convolution kernel. Thanks to the re-parameterization technique, the rotated kernels can be further merged into one single kernel with the original shape so that the computational cost remains the same. We further propose MOFA to provide auxiliary supervision for RKF. 
Experiments indicate that our method can improve the rotation invariance of feature descriptors remarkably on both HPatches and the DiverseBEV dataset.
\bibliographystyle{ieeetr} % We choose the "plain" reference style
\bibliography{references} % Entries are in the refs.bib file

\begin{thebibliography}{10}

\bibitem{schonberger2016structure}
J.~L. Schonberger and J.-M. Frahm, ``Structure-from-motion revisited,'' in {\em
  Proc. IEEE Conf. Comput. Vis. Pattern Recognit. (CVPR)}, pp.~4104--4113,
  2016.

\bibitem{mur2015orb}
R.~Mur-Artal, J.~M.~M. Montiel, and J.~D. Tardos, ``Orb-slam: a versatile and
  accurate monocular slam system,'' {\em IEEE Trans. Robot.}, vol.~31, no.~5,
  pp.~1147--1163, 2015.

\bibitem{lowe2004distinctive}
D.~G. Lowe, ``Distinctive image features from scale-invariant keypoints,'' {\em
  Int. J. Comput. Vis. (IJCV)}, vol.~60, no.~2, pp.~91--110, 2004.

\bibitem{bay2008speeded}
H.~Bay, A.~Ess, T.~Tuytelaars, and L.~Van~Gool, ``Speeded-up robust features
  (surf),'' {\em Comput. Vis. Image. Und.}, vol.~110, no.~3, pp.~346--359,
  2008.

\bibitem{rublee2011orb}
E.~Rublee, V.~Rabaud, K.~Konolige, and G.~Bradski, ``Orb: An efficient
  alternative to sift or surf,'' in {\em Proc. IEEE Int. Conf. Comput. Vision.
  (ICCV)}, pp.~2564--2571, Ieee, 2011.

\bibitem{yi2016lift}
K.~M. Yi, E.~Trulls, V.~Lepetit, and P.~Fua, ``Lift: Learned invariant feature
  transform,'' in {\em Proc. Eur. Conf. Comput. Vis. (ECCV)}, pp.~467--483,
  Springer, 2016.

\bibitem{ono2018lf}
Y.~Ono, E.~Trulls, P.~Fua, and K.~M. Yi, ``Lf-net: Learning local features from
  images,'' in {\em Proc. Adv. Neural Inf. Process. Syst. (NeurIPS)}, vol.~31,
  2018.

\bibitem{shen2019rf}
X.~Shen, C.~Wang, X.~Li, Z.~Yu, J.~Li, C.~Wen, M.~Cheng, and Z.~He, ``Rf-net:
  An end-to-end image matching network based on receptive field,'' in {\em
  Proc. Eur. Conf. Comput. Vis. (ECCV)}, pp.~8132--8140, 2019.

\bibitem{detone2018superpoint}
D.~DeTone, T.~Malisiewicz, and A.~Rabinovich, ``Superpoint: Self-supervised
  interest point detection and description,'' in {\em Proc. IEEE Int. Conf.
  Comput. Vision. (ICCV) Workshop}, pp.~224--236, 2018.

\bibitem{dusmanu2019d2}
M.~Dusmanu, I.~Rocco, T.~Pajdla, M.~Pollefeys, J.~Sivic, A.~Torii, and
  T.~Sattler, ``D2-net: A trainable cnn for joint description and detection of
  local features,'' in {\em Proc. IEEE Conf. Comput. Vis. Pattern Recognit.
  (CVPR)}, pp.~8092--8101, 2019.

\bibitem{revaud2019r2d2}
J.~Revaud, C.~De~Souza, M.~Humenberger, and P.~Weinzaepfel, ``R2d2: Reliable
  and repeatable detector and descriptor,'' in {\em Proc. Adv. Neural Inf.
  Process. Syst. (NeurIPS)}, vol.~32, 2019.

\bibitem{luo2020aslfeat}
Z.~Luo, L.~Zhou, X.~Bai, H.~Chen, J.~Zhang, Y.~Yao, S.~Li, T.~Fang, and
  L.~Quan, ``Aslfeat: Learning local features of accurate shape and
  localization,'' in {\em Proc. IEEE Conf. Comput. Vis. Pattern Recognit.
  (CVPR)}, pp.~6589--6598, 2020.

\bibitem{cohen2016group}
T.~Cohen and M.~Welling, ``Group equivariant convolutional networks,'' in {\em
  International conference on machine learning}, pp.~2990--2999, PMLR, 2016.

\bibitem{cohen2016steerable}
T.~S. Cohen and M.~Welling, ``Steerable cnns,'' {\em arXiv preprint
  arXiv:1612.08498}, 2016.

\bibitem{bokman2022case}
G.~B{\"o}kman and F.~Kahl, ``A case for using rotation invariant features in
  state of the art feature matchers,'' in {\em Proceedings of the IEEE/CVF
  Conference on Computer Vision and Pattern Recognition}, pp.~5110--5119, 2022.

\bibitem{parihar2021rord}
U.~S. Parihar, A.~Gujarathi, K.~Mehta, S.~Tourani, S.~Garg, M.~Milford, and
  K.~M. Krishna, ``Rord: Rotation-robust descriptors and orthographic views for
  local feature matching,'' in {\em 2021 IEEE/RSJ International Conference on
  Intelligent Robots and Systems (IROS)}, pp.~1593--1600, IEEE, 2021.

\bibitem{liu2019gift}
Y.~Liu, Z.~Shen, Z.~Lin, S.~Peng, H.~Bao, and X.~Zhou, ``Gift: Learning
  transformation-invariant dense visual descriptors via group cnns,'' {\em
  Advances in Neural Information Processing Systems}, vol.~32, 2019.

\bibitem{balntas2017hpatches}
V.~Balntas, K.~Lenc, A.~Vedaldi, and K.~Mikolajczyk, ``Hpatches: A benchmark
  and evaluation of handcrafted and learned local descriptors,'' in {\em Proc.
  IEEE Conf. Comput. Vis. Pattern Recognit. (CVPR)}, pp.~5173--5182, 2017.

\bibitem{rosten2006machine}
E.~Rosten and T.~Drummond, ``Machine learning for high-speed corner
  detection,'' in {\em Proc. Eur. Conf. Comput. Vis. (ECCV)}, pp.~430--443,
  Springer, 2006.

\bibitem{jaderberg2015spatial}
M.~Jaderberg, K.~Simonyan, A.~Zisserman, {\em et~al.}, ``Spatial transformer
  networks,'' in {\em Proc. Adv. Neural Inf. Process. Syst. (NeurIPS)},
  vol.~28, 2015.

\bibitem{sarlin2020superglue}
P.-E. Sarlin, D.~DeTone, T.~Malisiewicz, and A.~Rabinovich, ``Superglue:
  Learning feature matching with graph neural networks,'' in {\em Proc. IEEE
  Conf. Comput. Vis. Pattern Recognit. (CVPR)}, pp.~4938--4947, 2020.

\bibitem{sun2021loftr}
J.~Sun, Z.~Shen, Y.~Wang, H.~Bao, and X.~Zhou, ``Loftr: Detector-free local
  feature matching with transformers,'' in {\em Proc. IEEE Conf. Comput. Vis.
  Pattern Recognit. (CVPR)}, pp.~8922--8931, 2021.

\bibitem{weiler2019general}
M.~Weiler and G.~Cesa, ``General e (2)-equivariant steerable cnns,'' {\em
  Advances in Neural Information Processing Systems}, vol.~32, 2019.

\bibitem{peri2022ref}
A.~Peri, K.~Mehta, A.~Mishra, M.~Milford, S.~Garg, and K.~M. Krishna,
  ``Ref--rotation equivariant features for local feature matching,'' {\em arXiv
  preprint arXiv:2203.05206}, 2022.

\bibitem{szegedy2015going}
C.~Szegedy, W.~Liu, Y.~Jia, P.~Sermanet, S.~Reed, D.~Anguelov, D.~Erhan,
  V.~Vanhoucke, and A.~Rabinovich, ``Going deeper with convolutions,'' in {\em
  Proc. IEEE Conf. Comput. Vis. Pattern Recognit. (CVPR)}, pp.~1--9, 2015.

\bibitem{ding2021repvgg}
X.~Ding, X.~Zhang, N.~Ma, J.~Han, G.~Ding, and J.~Sun, ``Repvgg: Making
  vgg-style convnets great again,'' in {\em Proc. IEEE Conf. Comput. Vis.
  Pattern Recognit. (CVPR)}, pp.~13733--13742, 2021.

\bibitem{ding2022scaling}
X.~Ding, X.~Zhang, J.~Han, and G.~Ding, ``Scaling up your kernels to 31x31:
  Revisiting large kernel design in cnns,'' in {\em Proc. IEEE Conf. Comput.
  Vis. Pattern Recognit. (CVPR)}, pp.~11963--11975, 2022.

\bibitem{ding2019acnet}
X.~Ding, Y.~Guo, G.~Ding, and J.~Han, ``Acnet: Strengthening the kernel
  skeletons for powerful cnn via asymmetric convolution blocks,'' in {\em Proc.
  IEEE Int. Conf. Comput. Vision. (ICCV)}, pp.~1911--1920, 2019.

\bibitem{hinton2015distilling}
G.~Hinton, O.~Vinyals, J.~Dean, {\em et~al.}, ``Distilling the knowledge in a
  neural network,'' in {\em Proc. Adv. Neural Inf. Process. Syst. (NeurIPS)
  Workshop}, 2014.

\bibitem{li2017learning}
Z.~Li and D.~Hoiem, ``Learning without forgetting,'' {\em IEEE Trans. Pattern
  Anal. Mach. Intell. (TPAMI)}, vol.~40, no.~12, pp.~2935--2947, 2017.

\bibitem{peng2019few}
Z.~Peng, Z.~Li, J.~Zhang, Y.~Li, G.-J. Qi, and J.~Tang, ``Few-shot image
  recognition with knowledge transfer,'' in {\em Proc. IEEE Int. Conf. Comput.
  Vision. (ICCV)}, pp.~441--449, 2019.

\bibitem{chen2017learning}
G.~Chen, W.~Choi, X.~Yu, T.~Han, and M.~Chandraker, ``Learning efficient object
  detection models with knowledge distillation,'' in {\em Proc. Adv. Neural
  Inf. Process. Syst. (NeurIPS)}, vol.~30, 2017.

\bibitem{li2017mimicking}
Q.~Li, S.~Jin, and J.~Yan, ``Mimicking very efficient network for object
  detection,'' in {\em Proc. IEEE Conf. Comput. Vis. Pattern Recognit. (CVPR)},
  pp.~6356--6364, 2017.

\bibitem{he2019knowledge}
T.~He, C.~Shen, Z.~Tian, D.~Gong, C.~Sun, and Y.~Yan, ``Knowledge adaptation
  for efficient semantic segmentation,'' in {\em Proc. IEEE Conf. Comput. Vis.
  Pattern Recognit. (CVPR)}, pp.~578--587, 2019.

\bibitem{dou2020unpaired}
Q.~Dou, Q.~Liu, P.~A. Heng, and B.~Glocker, ``Unpaired multi-modal segmentation
  via knowledge distillation,'' {\em IEEE Trans. Pattern Anal. Mach. Intell.
  (TPAMI)}, vol.~39, no.~7, pp.~2415--2425, 2020.

\bibitem{2015FaceNet}
F.~Schroff, D.~Kalenichenko, and J.~Philbin, ``Facenet: A unified embedding for
  face recognition and clustering,'' in {\em Proc. IEEE Conf. Comput. Vis.
  Pattern Recognit. (CVPR)}, 2015.

\bibitem{2006Dimensionality}
R.~Hadsell, S.~Chopra, and Y.~Lecun, ``Dimensionality reduction by learning an
  invariant mapping,'' in {\em Proc. IEEE Conf. Comput. Vis. Pattern Recognit.
  (CVPR)}, 2006.

\bibitem{2016Real}
W.~Shi, J.~Caballero, F.~Huszár, J.~Totz, and Z.~Wang, ``Real-time single
  image and video super-resolution using an efficient sub-pixel convolutional
  neural network,'' in {\em Proc. IEEE Conf. Comput. Vis. Pattern Recognit.
  (CVPR)}, 2016.

\bibitem{shen2018mirror}
T.~Shen, Z.~Luo, L.~Zhou, R.~Zhang, S.~Zhu, T.~Fang, and L.~Quan, ``Matchable
  image retrieval by learning from surface reconstruction,'' in {\em Proc.
  Asian Conf. Comput. Vis. (ACCV)}, 2018.

\bibitem{schonberger2016pixelwise}
J.~L. Sch{\"o}nberger, E.~Zheng, J.-M. Frahm, and M.~Pollefeys, ``Pixelwise
  view selection for unstructured multi-view stereo,'' in {\em Computer
  Vision--ECCV 2016: 14th European Conference, Amsterdam, The Netherlands,
  October 11-14, 2016, Proceedings, Part III 14}, pp.~501--518, Springer, 2016.

\bibitem{mikolajczyk2005performance}
K.~Mikolajczyk and C.~Schmid, ``A performance evaluation of local
  descriptors,'' {\em IEEE transactions on pattern analysis and machine
  intelligence}, vol.~27, no.~10, pp.~1615--1630, 2005.

\bibitem{jin2021image}
Y.~Jin, D.~Mishkin, A.~Mishchuk, J.~Matas, P.~Fua, K.~M. Yi, and E.~Trulls,
  ``Image matching across wide baselines: From paper to practice,'' {\em
  International Journal of Computer Vision}, vol.~129, no.~2, pp.~517--547,
  2021.

\end{thebibliography}

% \end{sloppypar}
\end{document}